\PassOptionsToPackage{switch,mathlines}{lineno}

\documentclass[11pt]{article}

\usepackage[preprint]{acl}

\usepackage{times}
\usepackage{latexsym}
\usepackage[most]{tcolorbox}
\usepackage{xcolor}
\usepackage{multirow}
\usepackage{booktabs}
\usepackage{amsmath}
\usepackage{amssymb}
\usepackage{fvextra}

\usepackage[T1]{fontenc}

\usepackage[utf8]{inputenc}

\usepackage{microtype}

\usepackage{inconsolata}

\usepackage{graphicx}

\title{EvoGraph-Mem: Failure-Aware Editable Graph Memory for Long-Term Language Agents
}

\author{
  \textbf{Yuxi Qian} \\
  \And
  \textbf{Yuxiang Ren\textsuperscript{*}} \\
}

\begin{document}
\maketitle
\begin{abstract}
Long-term memory is essential for language agents operating across extended interactions and evolving tasks. Existing memory-augmented agents mainly focus on storing and retrieving past experience, but the quality of stored memories may degrade over time. In particular, previously distilled insights can become outdated, over-generalized, or harmful under new task contexts, causing memory pollution when repeatedly reused. To address this issue, we study insight-level memory maintenance for long-term language agents and propose a failure-aware memory maintenance framework based on an editable insight graph. Each insight node tracks positive evidence, negative evidence, and an activation state, enabling the agent to distinguish reusable insights from conflicting or invalid ones. We further introduce a utility-aware retrieval mechanism and a graph controller that updates the memory graph after task execution by keeping reliable insights, archiving invalid ones, revising outdated ones, and adding newly discovered reusable insights. Extensive experiments show that our method consistently outperforms representative memory-based agent baselines across different backbone models. Ablation studies further demonstrate that append-only memory is insufficient for long-horizon tasks, while evidence-aware retrieval and graph-level editing improve memory reliability and downstream task performance.
\end{abstract}

\section{Introduction}

Large language model (LLM) agents are increasingly expected to operate across extended interactions, multi-step tasks, and evolving environments. Long-term memory is therefore essential for retaining useful experience and reusing it in future decision making. Existing memory-augmented agents store past interactions, reflections, or reusable skills to improve long-horizon behavior \citep{zhong2024memorybank,wang2023voyager}. However, long-term memory is not only a problem of storage and retrieval, but also one of maintaining memory quality over time.

As agents encounter diverse tasks, previously distilled insights may become outdated, over-generalized, or harmful under new contexts. Once such insights are repeatedly retrieved, they may introduce memory pollution and degrade downstream reasoning. This issue is especially important for high-level insights, which are intended to generalize across tasks but may cause greater harm when applied beyond their valid contexts. Recent graph-based memory systems, such as G-Memory, improve memory organization by structuring historical experience into insight, query, and interaction graphs \citep{zhang2025gmemory}. Nevertheless, existing memory systems remain limited in explicitly modeling whether a recalled insight was helpful, conflicting, or should be revised after task failure.

In this paper, we study \emph{insight-level memory maintenance} for long-term language agents and propose a failure-aware memory maintenance framework based on an editable insight graph. Each insight node explicitly tracks positive evidence, negative evidence, and an activation state: positive evidence records queries for which the insight provides useful guidance, negative evidence captures cases where the insight is ineffective or harmful, and the activation state distinguishes active insights from archived ones to prevent invalid memories from being repeatedly retrieved. Building on this representation, we introduce utility-aware retrieval and a graph controller for memory correction. Candidate insights are ranked by jointly considering positive support and conflicting evidence, while after task execution, the graph controller updates the insight graph by keeping reliable insights, archiving invalid ones, revising outdated ones, and adding new insights when reusable knowledge is discovered. Unlike model editing methods that modify internal parameters \citep{mitchell2021fast}, our method operates at the external agent-memory level and keeps the underlying LLM unchanged.

Extensive experiments show that our framework consistently improves over representative memory-based agent methods. Ablation studies further demonstrate that append-only memory is insufficient, while evidence-aware retrieval and graph-level editing contribute to stronger long-horizon performance.

Our contributions are summarized as follows:
\begin{itemize}
    \item Identify insight-level memory maintenance as a challenge for long-term language agents, where useful insights may later become over-generalized, conflicting, or harmful.
    \item Propose an editable insight graph that tracks positive evidence, negative evidence, and activation states for failure-aware retrieval.
    \item Introduce a graph controller that performs corrective memory updates through keeping, archiving, revising, and adding insights.
    \item Experiments demonstrate consistent improvements over representative memory-based agent baselines, with ablations validating the importance of graph-level editing.
\end{itemize}

\begin{figure*}[t]
  \includegraphics[width=\linewidth]{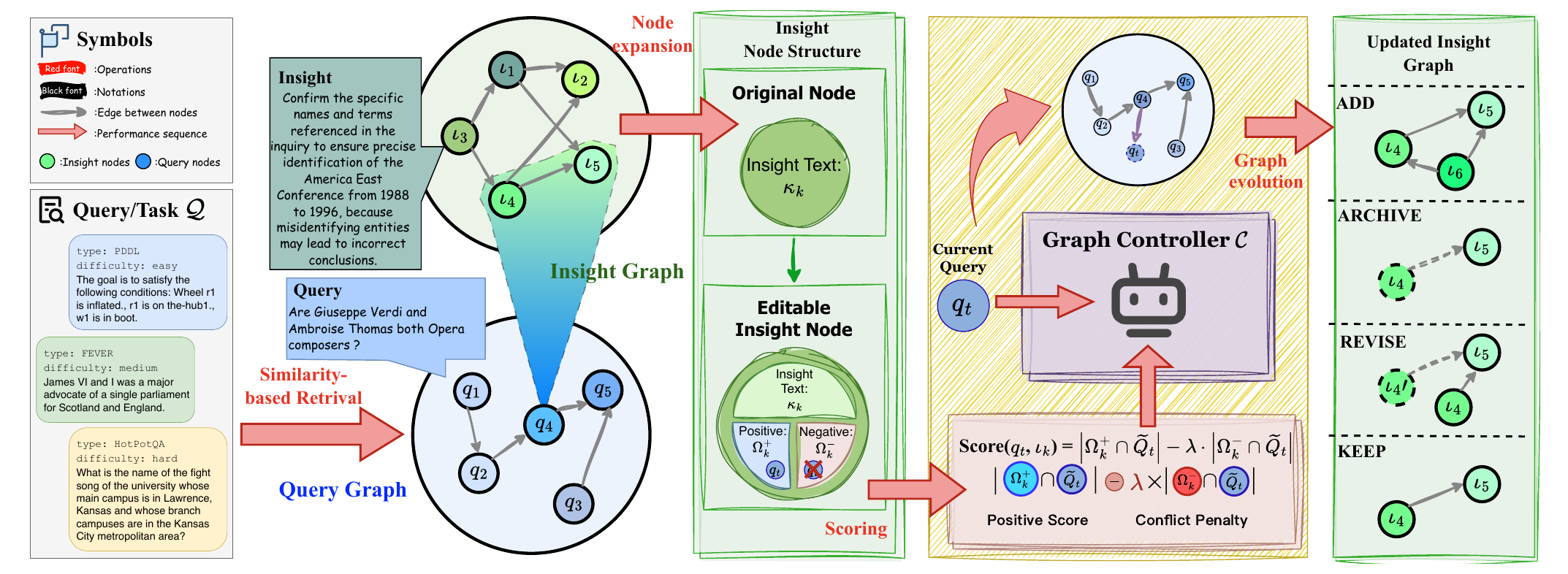}
    \caption{The mechanism of memory graph update}
    \label{fig:arch}
\end{figure*}

\section{Related Work}
\label{sec:related}

\subsection{Long-Term Memory for LLM Agents}

Long-term memory has become a key component for language agents that operate across extended interactions, tasks, or environments. Existing systems represent persistent experience in different forms, including conversational histories, episodic memories, reflective summaries, and executable skills. Generative Agents maintain memory streams and synthesize higher-level reflections to support coherent interactive behavior \citep{park2023generative}. Reflexion converts task feedback into verbal reflections stored in episodic memory, enabling agents to improve later decisions without parameter updates \citep{shinn2023reflexion}. MemoryBank studies long-term conversational memory for recalling and updating past interactions \citep{zhong2024memorybank}, while Voyager accumulates reusable skills for open-ended embodied exploration \citep{wang2023voyager}. MemGPT further frames memory management as virtual context management across memory tiers \citep{packer2023memgpt}. Recent systems such as MemoryOS, Mem0, and A-MEM move beyond simple storage by emphasizing hierarchical organization, dynamic updating, scalable retrieval, and adaptive memory networks \citep{kang2025memory,chhikara2025mem0,xu2026mem}. These studies show that persistent memory is essential for long-horizon agents. However, they largely focus on how memories are stored, retrieved, summarized, or expanded, while the validity of a distilled insight after repeated reuse remains less explicitly modeled. Our work complements this direction by focusing on insight-level memory maintenance, where retrieved memories are evaluated after task execution and may be reinforced, suppressed, or revised.

\subsection{Graph Memory and Corrective Knowledge Maintenance}

Graph structures provide a natural mechanism for organizing relational dependencies among tasks, experiences, and abstract knowledge. In retrieval-augmented generation, HippoRAG combines language models, knowledge graphs, and graph-based retrieval to support long-term knowledge integration and multi-hop reasoning \citep{gutierrez2024hipporag}. For agent memory, AriGraph integrates semantic and episodic memories into a graph-based world model \citep{anokhin2024arigraph}, and Zep introduces a temporal knowledge graph architecture for continuously evolving conversational memory \citep{rasmussen2025zep}. Closely related to our work, G-Memory organizes multi-agent histories into a hierarchical graph consisting of insight, query, and interaction graphs, enabling agents to retrieve both high-level insights and fine-grained interaction trajectories \citep{zhang2025gmemory}. These graph-based approaches improve memory organization and retrieval, but they remain limited in explicitly handling cases where a previously useful insight becomes over-generalized, conflicting, or harmful under new task contexts.

Our work is closely related to knowledge editing, which seeks to correct outdated or erroneous knowledge in language models without full retraining. Existing approaches, such as MEND, ROME, MEMIT, and GRACE, modify model parameters or auxiliary memory structures to update factual associations while preserving the model's general behavior \citep{mitchell2021fast,meng2022locating,meng2022mass,hartvigsen2023aging}. Subsequent methods, including AlphaEdit and AdaEdit, further examine how edited knowledge can be preserved under sequential or continuous editing scenarios \citep{fang2025alphaedit,li-chu-2025-adaedit}. Despite this shared objective, our work differs in that it addresses unreliable knowledge at the external agent-memory level, rather than through parameter-level intervention. We introduce an editable insight graph in which nodes record positive evidence, negative evidence, and activation states. This structure enables the system to archive invalid insights, update outdated ones, and mitigate memory pollution during long-term agent deployment.


\section{Methodology}
This section presents our failure-aware memory maintenance framework for long-term agent memory, as illustrated in Figure~\ref{fig:arch}. Given a new user query $Q$, the agent first retrieves relevant historical 
queries and insights following the graph-based memory structure of G-memory 
\citep{zhang2025gmemory}. However, unlike append-only memory systems, our framework 
explicitly models whether a recalled insight remains useful, becomes harmful, or 
requires revision after being applied to a new task. 

Specifically, we extend each insight node with positive evidence, negative evidence, 
and an activation state, enabling the system to distinguish reusable insights from 
non-generalizable or polluted ones. After task completion, a graph controller updates 
the insight graph through three maintenance operations: keeping reliable insights, 
archiving invalid insights, and revising outdated insights. This design allows the 
memory graph to evolve from raw accumulation toward corrective maintenance, thereby 
reducing the long-term propagation of invalid or harmful memories.

\subsection{Graph-Based Historical Memory}
\label{sec: tf}

During the initial forward pass, the query graph is defined as:
\begin{equation}
\mathcal{G}_{\text{query}} = (\mathcal{Q}, \mathcal{E}_q) 
= \left( \left\{ Q_i, \Psi_i, G^{(Q_i)}_{\text{inter}} \right\}_{i=1}^{|\mathcal{Q}|}, \mathcal{E}_q \right).
\end{equation}
Here, $\mathcal{Q} = \{q_i\}$ denotes the set of query nodes, where each node $q_i \triangleq (Q_i, \Psi_i, G^{(Q_i)}_{\text{inter}})$ consists of the original query $Q_i$, its task status $\Psi_i \in \{\text{Failed}, \text{Resolved}\}$, and the corresponding interaction graph $G^{(Q_i)}_{\text{inter}}$. The edge set $\mathcal{E}_q \subseteq \mathcal{Q} \times \mathcal{Q}$ represents semantic dependencies between queries. Leveraging this structured topology, the query graph supports more precise retrieval compared to coarse similarity-based approaches such as embedding matching.

Upon receiving a new query $Q$, the multi-agent system performs a retrieval operation on the graph $\mathcal{G}_{\text{query}}$ maintained from historical tasks to identify and recall analogous instances: 
\begin{equation}
    \mathcal{Q}^{\mathcal{S}}=\underset{q_{i} \in \mathcal{Q} \text { s.t. }\left|\mathcal{Q}^{\mathcal{S}}\right|=k}{\arg \text { top-k }}\left(\frac{\mathbf{v}(Q) \cdot \mathbf{v}\left(q_{i}\right)}{|\mathbf{v}(Q)|\left|\mathbf{v}\left(q_{i}\right)\right|}\right).
\end{equation}
In this formulation, $\mathbf{v}(\cdot)$ denotes the vectorization applied to both the historical queries and the current query. By computing the cosine similarity between these representations, the system can retrieve historical queries that exhibit high semantic similarity to the current one. However, relying exclusively on this metric may introduce noise. To mitigate this, corresponding extension techniques are further incorporated:
\begin{equation}
  \Delta \mathcal{Q} = \{{Q}_{k} \in \mathcal{Q} \mid \exists Q_{j} \in \mathcal{Q}^{\mathcal{S}} \text{ s.t. } Q_{k} \in \mathcal{N}(Q_{j})\}.
\end{equation}
\begin{equation}
  \tilde{\mathcal{Q}}^{\mathcal{S}} = \mathcal{Q}^{\mathcal{S}} \cup \Delta \mathcal{Q}.
\end{equation}
Here, $\Delta \mathcal{Q}$ denotes the set of neighboring nodes for all nodes within $\mathcal{Q}^{\mathcal{S}}$. Through the expansion of the original $\tilde{\mathcal{Q}}^{\mathcal{S}}$, a task-specific subgraph is subsequently constructed based on these retrieved elements.

The insight graph $\mathcal{G}_{\text{insight}}$ is defined as:
\begin{equation}
\mathcal{G}_{\text{insight}} = (\mathcal{I}, \mathcal{E}_{\text{i}}) = \left( \left\langle \kappa_k, \Omega_k \right\rangle_{k=1}^{|\mathcal{I}|}, \mathcal{E}_{\text{i}} \right).
\end{equation}
Here, the node set $\mathcal{I} = \{\iota_k\}$ denotes the distilled insights, with each node $\iota_k$ comprising the insight content $\kappa_k$ and an associated set of supporting queries $\Omega_k \subseteq \mathcal{Q}$. Furthermore, the edge set $\mathcal{E}_{\text{i}} \subseteq \mathcal{I} \times \mathcal{I} \times \mathcal{Q}$ establishes hyper-connections, wherein a tuple $(\iota_m, \iota_n, q_j)$ indicates that insight $\iota_m$ contextualizes $\iota_n$, mediated by query $q_j$.
\subsection{Failure-Aware Insight Representation}
\label{sec: ne}

A central limitation of existing insight memories is that they typically store only 
the queries that contributed to the creation of an insight. Such a representation 
implicitly assumes that an insight remains valid once generated. In long-term agent 
deployment, however, this assumption can lead to memory pollution: an insight that 
was useful for one query may be repeatedly retrieved for semantically similar but 
incompatible tasks, thereby degrading future reasoning. 

To make insight validity explicitly controllable, we represent each insight node as
\begin{equation}
    \iota_k = (\kappa_k, \Omega_k^{+}, \Omega_k^{-}, z_k),
\end{equation}
where $\kappa_k$ denotes the insight content, $\Omega_k^{+}$ denotes the set of 
queries for which the insight provides positive support, $\Omega_k^{-}$ denotes the 
set of queries for which the insight is ineffective or harmful, and 
$z_k \in \{0,1\}$ denotes the activation state of the node. An active node 
($z_k=1$) can be retrieved for future memory augmentation, whereas an archived node 
($z_k=0$) is excluded from retrieval but retained for auditability and edit tracing.

\subsection{Utility-Aware Insight Retrieval}
\label{sec:retrieval}

Given the expanded query subgraph $\tilde{\mathcal{Q}}^{\mathcal{S}}$, we retrieve 
insights that are both active and supported by at least one structurally related 
historical query:
\begin{equation}
    I_t^{\mathrm{cand}} =
    \left\{
    \iota_k
    \mid
    z_k = 1,\;
    \Omega_k^{+} \cap \tilde{\mathcal{Q}}^{\mathcal{S}} \neq \varnothing
    \right\}.
\end{equation}

However, positive relevance alone is insufficient: an insight may be useful for some 
historical queries while conflicting with others. To suppress polluted or 
non-generalizable insights, we score each candidate using both positive and negative 
evidence:
\begin{equation}
    r_{t,k}=s_{t,k}-\lambda c_{t,k},
\end{equation}
where
\begin{equation}
    s_{t,k}=|\Omega_k^{+}\cap \tilde{\mathcal{Q}}^{\mathcal{S}}|,
    \quad
    c_{t,k}=|\Omega_k^{-}\cap \tilde{\mathcal{Q}}^{\mathcal{S}}|.
\end{equation}
Here, $s_{t,k}$ measures the amount of relevant positive evidence, while $c_{t,k}$ 
measures the amount of conflicting historical evidence. The coefficient 
$\lambda \ge 0$ controls the strength of conflict penalization. We then select the 
top-$B$ insights:
\begin{equation}
    I_t^{\mathrm{ret}}
    =
    \operatorname{TopB}_{\iota_k \in I_t^{\mathrm{cand}}}
    r_{t,k}.
\end{equation}

\subsection{Graph Controller for Memory Correction}
\label{sec:gc}

After the agent completes the current task, the retrieved insights are no longer 
treated as static memories. Instead, the graph controller evaluates whether each 
insight retrieved was helpful, harmful, or insufficient for the current query. This 
post-task feedback enables corrective memory maintenance.
We instantiate the graph controller as a prompt-constrained LLM. Given the current query, execution trajectory, task feedback, and retrieved insight nodes, the controller is required to output a valid object whose operations are restricted to KEEP, ARCHIVE, and REVISE for existing insights, with a separate binary ADD decision for new insight creation.

For each retrieved insight $\iota_k \in I_t^{\mathrm{ret}}$, the controller predicts 
an edit operation:
\begin{equation}
\begin{aligned}
        o_k& = \mathcal{C}_{\mathrm{edit}}(Q, \Psi,I_t^{\mathrm{ret}}, \iota_k),\\
      o_k \in& \{\text{Keep}, \text{Archive}, \text{Revise}\},
\end{aligned}
\end{equation}
where $\Psi$ denotes the final task feedback. In addition, the controller separately 
decides whether the current task yields a new reusable insight $\hat{\kappa}$:
\begin{equation}
    a = \mathcal{C}_{\mathrm{add}}(Q, \Psi, I_t^{\mathrm{ret}}, \hat{\kappa}),
    \quad
    a \in \{0,1\}.
\end{equation}
This separation avoids conflating the correction of existing insights with the 
creation of new memory nodes.
\paragraph{Archive.}
If an insight is judged to be invalid or harmful for the current task, we deactivate 
the node and record the current query as negative evidence:
\begin{equation}
    (\kappa_k,\Omega_k^{+},\Omega_k^{-},z_k)
    \leftarrow
    (\kappa_k,\Omega_k^{+},\Omega_k^{-}\cup\{q\},0).
\end{equation}
The archived node is excluded from future retrieval but retained in the graph to 
preserve the failure evidence and support edit traceability.

\paragraph{Revise.}
If an insight is partially useful but over-generalized or outdated, the controller 
archives the old node and creates a revised node. The old node is updated as
\begin{equation}
    (\kappa_k,\Omega_k^{+},\Omega_k^{-},z_k)
    \leftarrow
    (\kappa_k,\Omega_k^{+},\Omega_k^{-}\cup\{q\},0).
\end{equation}
Then, a revised insight node is created:
\begin{equation}
    \iota_{k'}
    =
    (\kappa'_{k},\Omega_k^{+}\cup\{q\},\varnothing,1).
\end{equation}
This operation makes the correction trace explicit: the old insight is preserved as 
an invalidated version, while the revised insight is activated as a corrected memory 
for future tasks.

\paragraph{Add.}
If the current task reveals a reusable pattern that is not covered by existing 
retrieved insights, the controller creates a new insight node:
\begin{equation}
    \iota_{\mathrm{new}}
    =
    (\hat{\kappa}, \{q\}, \varnothing, 1).
\end{equation}
Here, $\hat{\kappa}$ denotes the newly distilled insight and the current query $q$ 
serves as its initial positive evidence. This operation allows the memory graph to 
expand only when the current task contributes novel reusable knowledge.

\paragraph{Keep.}
If a retrieved insight contributes positively to the current task, the controller 
preserves the insight and strengthens its positive evidence:
\begin{equation}
    (\kappa_k,\Omega_k^{+},\Omega_k^{-},z_k)
    \leftarrow
    (\kappa_k,\Omega_k^{+}\cup\{q\},\Omega_k^{-},1).
\end{equation}
Repeated positive evidence increases the likelihood that the insight will be 
retrieved for future structurally related tasks.

\paragraph{Edge Update.}
In addition to node-level maintenance, the controller updates the insight graph 
connectivity. When a new or revised insight is created, we connect it with the 
retrieved insights that jointly contributed to the current task:
\begin{equation}
\begin{aligned}
    \mathcal{E}_{i}
    \leftarrow &
    \mathcal{E}_{i}
    \cup
    \{(\iota_a,\iota_b,q)
    \mid
    \iota_a \in I_t^{\mathrm{ret}}, \\
    \iota_b& \in \mathcal{I}_{\mathrm{new}},
    \Psi=\text{Resolved}\},
\end{aligned}
\end{equation}
where $\mathcal{I}_{\mathrm{new}}$ denotes the set of newly added or revised active 
insight nodes. For archived nodes, existing edges are retained for traceability but 
ignored during active retrieval. This update allows the graph to encode not only 
which insights are valid, but also how corrected insights emerge from prior memory 
contexts.

\begin{table*}[t]
\centering
\small
\begin{tabular}{l|cccc|cccc}
\toprule
\multirow[t]{2}{*}{Model} & \multicolumn{4}{c|}{GPT-4o-mini} & \multicolumn{4}{c}{Qwen2.5-7B} \\
\cmidrule(lr){1-1} \cmidrule(lr){2-5} \cmidrule(lr){6-9}
Method & PDDL & HotpotQA & FEVER & Avg. & PDDL & HotpotQA & FEVER & Avg. \\
\midrule
None        & 23.53 & 28.57 & 57.13 & 36.41 & 16.17 & 33.33 & 58.74 & 36.08 \\
MemoryBank  & 20.41 & 33.67 & 61.22 & 38.43 & 14.83 & 32.67 & 59.45 & 35.65 \\
Voyager     & 24.56 & 32.32 & 63.27 & 40.05 & 12.00 & 34.29 & 52.44 & 32.91 \\
Generative  & 25.53 & 31.63 & 60.20 & 39.12 & 17.88 & 34.17 & 61.25 & 37.77 \\
G-Memory    & 27.77 & 35.67 & 66.24 & 43.23 & 21.01 & 37.34 & 64.34 & 40.90 \\
\textbf{Ours} & \textbf{30.67} & \textbf{43.43} & \textbf{68.32} & \textbf{47.47} 
              & \textbf{23.40} & \textbf{39.22} & \textbf{65.35} & \textbf{42.66} \\
\midrule
Improvement (\%) 
& 10.44 & 21.75 & 3.14 & 9.81 
& 11.38 & 5.03 & 1.57 & 4.30 \\
\bottomrule
\end{tabular}
\caption{Performance comparison of different memory methods on PDDL, HotpotQA, and FEVER datasets.}
\label{tab:memory_comparison}
\end{table*}

\begin{figure*}[t]
  \includegraphics[width=\linewidth]{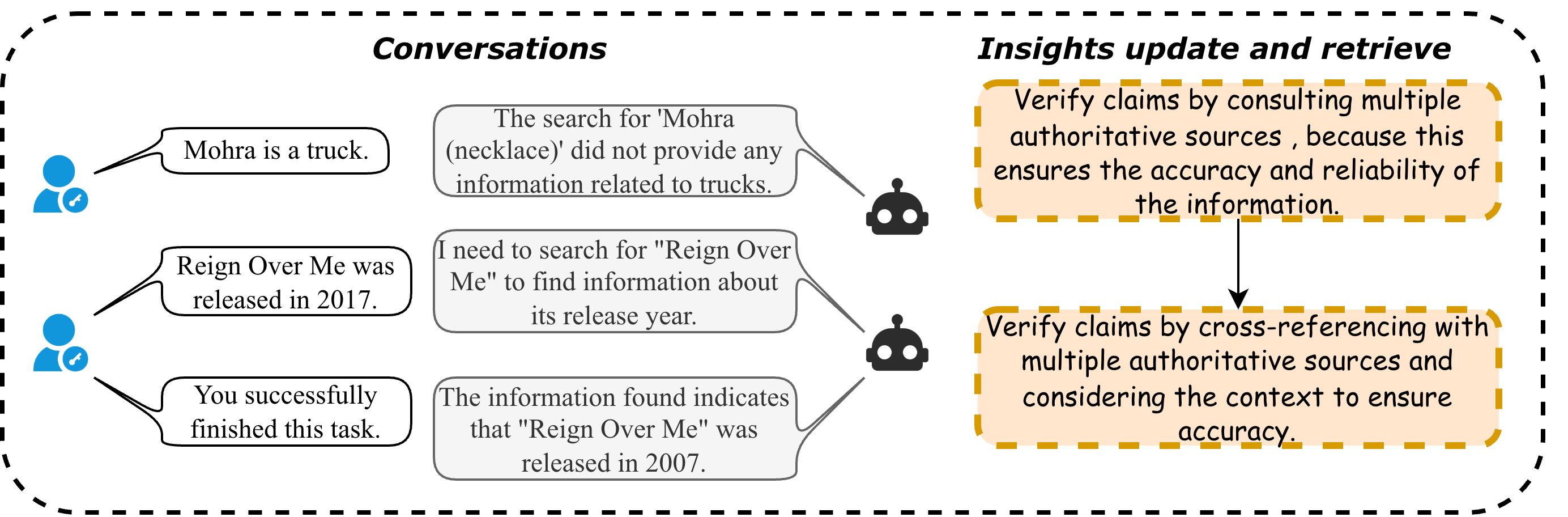}
    \caption{Case study demonstrating the evolution of insights upon FEVER tasks} 
    \label{fig:case}
\end{figure*}

\begin{figure}[t]
\centering
\includegraphics[width=\columnwidth]{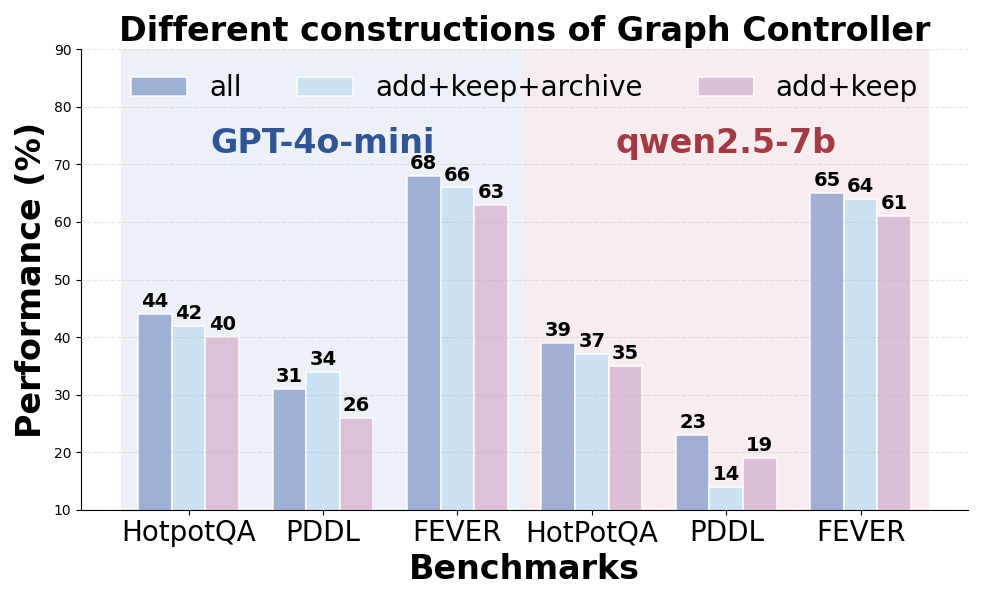}
\caption{Ablation results on the benchmarks with GPT-4o-mini and qwen2.5-7b}
\label{fig:ablation}
\end{figure}

\begin{table}[t]
\centering
\small
\setlength{\tabcolsep}{4pt}
\begin{tabular}{lcccc}
\toprule
\multirow{2}{*}{Method} 
& \multicolumn{2}{c}{GPT-4o-mini} 
& \multicolumn{2}{c}{Qwen2.5-7B} \\
\cmidrule(lr){2-3} \cmidrule(lr){4-5}
& Tokens & $\Delta_{\text{None}}$ 
& Tokens & $\Delta_{\text{None}}$ \\
\midrule
None        & 3.7M & 0.0\%   & 3.3M & 0.0\%   \\
MemoryBank  & 4.7M & +27.0\% & 3.5M & +6.1\%  \\
Voyager     & 5.5M & +48.6\% & 4.9M & +48.5\% \\
Generative  & 6.0M & +62.2\% & 4.3M & +30.3\% \\
G-Memory    & 6.2M & +67.6\% & 5.3M & +60.6\% \\
Ours        & 6.4M & +73.0\% & 5.7M & +72.7\% \\
\bottomrule
\end{tabular}
\caption{
Average token consumption across PDDL, HotpotQA, and FEVER.
$\Delta_{\text{None}}$ denotes the relative token change compared with the no-memory baseline.
}
\label{tab:token_consumption}
\end{table}

\section{Experiments}
\subsection{Experiment Setup}
\paragraph{Datasets.}
We conduct our experiments on PDDL \citep{ma2024agentboard}, HotpotQA \cite{yang2018hotpotqa} and FEVER \cite{thorne2018fever}.The PDDL dataset is a collection of strategic games—including Gripper, Barman, Blocksworld, and Tyreworld—adapted to evaluate agents in multi-turn scenarios. It requires multiple rounds of planned actions to finish a single subgoal, testing the agent's ability to plan strategically and avoid repetitive steps. The HotpotQA dataset challenges systems to find and reason over multiple supporting documents to arrive at an answer, providing sentence-level supporting facts to facilitate explainable predictions. The FEVER dataset is a large-scale benchmark for fact extraction and verification. It evaluates a model's ability to retrieve necessary textual evidence and classify claims into \textsc{Supported}, \textsc{Refuted} or \textsc{NotEnoughInfo} categories.

\paragraph{Evaluation Metrics.}
For the FEVER and HotpotQA datasets, we use \textit{exact match} accuracy. For the PDDL datasets, we use \textit{the progress rate} to measure the progress of a task.

\paragraph{Compared Methods.}
We compare ours with several representative memory methods, including:

\paragraph{MemoryBank \cite{zhong2024memorybank} :}
The framework is a long-term memory mechanism tailored for Large Language Models. It enables AI to store, recall, and update past interactions. Inspired by the Ebbinghaus Forgetting Curve, it continually adapts to a user's personality.

\paragraph{Voyager \cite{wang2023voyager} :}
This method is the first LLM-powered embodied lifelong learning agent in Minecraft that continuously explores the world, acquires diverse skills, and makes novel discoveries without human intervention. It features an automatic curriculum, a skill library, and an iterative prompting mechanism.

\paragraph{Generative Agents \cite{park2023generative} :}
It operates on a system comprising foundational observational records and higher-order cognitive reflections. By deducing patterns and logically synthesizing fragmented interactions, the reflection layer distills abstract beliefs—ultimately elevating raw experiences into a highly structured and conceptually deep knowledge representation.

\paragraph{G-Memory \cite{zhang2025gmemory} :}
This framework is a hierarchical graph memory architecture. It manages lengthy collaboration histories through Insight, Query, and Interaction graph layers, enabling agents to efficiently retrieve past experiences for continuous team learning and self-evolution.

\subsection{Main Results}
The experimental results in PDDL, HotpotQA and FEVER datasets are shown in Table~\ref{tab:memory_comparison}. We summarize the key observations as follows: 

(1) Among the evaluated memory methods, MemoryBank and Voyager demonstrate relatively weaker performance across the benchmarks. This indicates that simply relying on the Ebbinghaus Forgetting Curve for memory decay or maintaining a fixed skill library  is insufficient for dynamically managing complex, multi-turn task contexts. Generative Agents attempts to bridge this gap by synthesizing fragmented interactions into higher-order cognitive reflections, however, its unstructured reflection mechanism struggles to maintain precise semantic dependencies, leading to suboptimal reasoning. G-Memory significantly outperforms these baselines by organizing lengthy histories into a structured, hierarchical graph architecture, yet it lacks a systematic memory management mechanism. G-Memory operates on a conventional append-only design, this inevitably causes memory overload and introduces noise from non-generalizable or obsolete insights.

(2) Our failure-aware memory maintenance framework resolves these issues by introducing a dynamic Graph Controller and an extended Insight Node structure. Instead of passively accumulating data, our revised version supports active memory manipulation—such as modifying obsolete memories and pruning invalid representations. To prevent interference of different queries, we partition the supporting queries into positive ($\Omega_k^+$) and negative ($\Omega_k^-$) subsets , utilizing a conflict penalty mechanism to dynamically score and rank memory utility. This ensures that only the most task-aligned, highly generalizable insights are retrieved while mitigating the adverse impact of irrelevant data. By explicitly addressing the structural flaws of prior methods, our approach yields a 10.44\% and 11.38\% progress rate improvement on the multi-turn PDDL dataset, alongside a 21.75\% and 5.03\% exact match accuracy enhancement on the complex HotpotQA dataset across the two respective models.

(3) We further analyze the token consumption of different memory methods in Table~\ref{tab:token_consumption}. As expected, memory-augmented methods generally consume more tokens than the no-memory baseline, since they introduce additional retrieval, summarization, or memory-update steps. Among them, our method incurs the highest overall token usage, with a relative increase of 73.0\% on GPT-4o-mini and 72.7\% on Qwen2.5-7B compared with the no-memory setting. This overhead mainly comes from evidence-aware insight retrieval and graph-controller-based memory correction. However, compared with G-Memory, which is the strongest graph-based baseline, the additional token cost of our method is relatively moderate: token usage increases from 6.2M to 6.4M on GPT-4o-mini and from 5.3M to 5.7M on Qwen2.5-7B. Considering the consistent performance gains reported in Table~\ref{tab:memory_comparison}, these results suggest that our framework trades a limited amount of additional computation for more reliable memory utilization. These results indicate a performance-cost trade-off: while corrective memory maintenance introduces additional token overhead, it also improves the reliability and task utility of long-term memory.

\subsection{Ablation Study}

To evaluate the contribution of different operations in the proposed graph controller, we compare three variants: 
\textit{add+keep}, which only appends new insights and retains existing ones; 
\textit{add+keep+archive}, which further allows the controller to deactivate outdated or harmful insights; 
and \textit{all}, which uses the complete graph-controller design.

As shown in Figure~\ref{fig:ablation}, the complete controller generally achieves the strongest performance across both backbone models. 
On GPT-4o-mini, the full variant obtains 44\%, 31\%, and 68\% on HotpotQA, PDDL, and FEVER, respectively. 
Compared with the simple \textit{add+keep} variant, this corresponds to gains of 4 points on HotpotQA, 5 points on PDDL, and 5 points on FEVER. 
A similar trend can be observed on Qwen2.5-7B, where the full controller achieves 39\%, 23\%, and 65\%, outperforming \textit{add+keep} by 4, 4, and 4 points on the three benchmarks, respectively.

The comparison between \textit{add+keep} and \textit{add+keep+archive} further demonstrates the importance of memory deactivation. 
Adding the archive operation improves performance in most settings, especially on GPT-4o-mini PDDL, where the score increases from 26\% to 34\%, and on FEVER, where it improves from 63\% to 66\%. 
This suggests that long-term memory should not be treated as a purely append-only repository: obsolete or misleading insights can introduce noise into retrieval and harm downstream decision making.

\subsection{Case Study}
To visually demonstrate whether the memory within the graph is effectively updated, we extract the system's insight retrieval capabilities, alongside the explicit content of the recalled insights, across two consecutive tasks. We present case studies in Figure~\ref{fig:case}. As illustrated in the figure, upon receiving the initial task "Mohra is a truck", the system retrieves the insight "Verify claims by consulting multiple authoritative sources..." from memory to guide the ongoing task execution. Subsequently, during the execution of the second task, the recalled insight transitions to "Verify claims by cross-referencing with multiple authoritative sources and considering the context...". This progression demonstrates that the insights undergo dynamic self-evolution and refinement throughout the sequence of tasks. Consequently, these adjusted insights become increasingly fine-grained and better adapted to novel requirements, playing a pivotal role in facilitating ultimate task success.

\section{Conclusion}
\label{sec:conclusion}

In this paper, we studied insight-level memory maintenance for long-term language agents, where previously distilled insights may become ineffective, over-generalized, or harmful under new task contexts. We proposed a failure-aware memory maintenance framework that extends graph-based agent memory with editable insight nodes. Each insight tracks positive evidence, negative evidence, and an activation state, while a graph controller updates the memory graph through keeping, archiving, revising, and adding insights after task execution.

Experiments on PDDL, HotpotQA, and FEVER show that our method improves over representative memory-based agent frameworks across different backbone models. The ablation study further suggests that append-only memory is insufficient for long-horizon tasks, and that evidence-aware retrieval together with graph-level editing contributes to stronger performance. Overall, our work shifts long-term agent memory from passive accumulation toward corrective maintenance.

\section{Limitations}
\label{sec:limitations}

Our framework has several limitations. First, the graph controller depends on post-task feedback, and noisy or incomplete feedback may lead to suboptimal edits, such as archiving useful insights or revising memories prematurely. Second, the positive and negative evidence sets provide a simple validity signal but do not fully capture the context or degree of an insight's usefulness. Third, our experiments focus on PDDL, HotpotQA, and FEVER, leaving broader long-term agent scenarios, such as web interaction, software engineering, and personalized assistants, for future evaluation. Finally, maintaining and editing an insight graph introduces additional computational and storage overhead, which may require more efficient pruning and controller invocation strategies in large-scale deployments.

\bibliography{custom}

\clearpage
\appendix
\renewcommand{\thesection}{Appendix \Alph{section}}

\section{prompt-constrained Graph Controller}

This section presents the prompt used to instantiate the prompt-constrained
graph controller in our memory maintenance framework. Given the current query,
the agent trajectory, the final answer, task feedback, and the retrieved
insight nodes, the controller is required to output structured memory edit
decisions. Specifically, it selects one operation from \textsc{Keep},
\textsc{Archive}, and \textsc{Revise} for each retrieved insight, and separately
determines whether a new reusable insight should be added to the memory graph.
The output is constrained to a predefined JSON format, which enables stable and
automatic updates of positive evidence, negative evidence, and activation states
in the editable insight graph.

\begin{tcolorbox}[
    enhanced,
    breakable,
    colback=gray!3,
    colframe=black!75,
    coltitle=white,
    title=\textbf{Graph Controller},
    fonttitle=\bfseries\large,
    attach boxed title to top left={xshift=0mm,yshift=0mm},
    boxed title style={
        colback=black!80,
        colframe=black!80,
        boxrule=0pt,
        sharp corners
    },
    sharp corners,
    boxrule=0.8pt,
    left=8pt,
    right=8pt,
    top=8pt,
    bottom=8pt
]
You are a graph memory controller. Update retrieved insight nodes after a task.

For each retrieved insight, choose exactly one operation:

KEEP:
Choose this if the insight was valid and directly useful for the current task.
The current query will be added to positive evidence.

ARCHIVE:
Choose this if the insight was invalid, harmful, misleading, or contradicted by the current task.
The current query will be added to negative evidence and the node will be deactivated.

REVISE:
Choose this if the insight was partially useful but too broad, outdated, or missing necessary conditions.
The old node will be archived, and a new revised active node will be created.
The revised insight must correct or narrow the original insight, not merely paraphrase it.

ADD:
After editing existing insights, decide whether the current task contains a new reusable insight not covered by retrieved insights.
Add only generalizable and non-duplicated insights.

Use only the given task information. Be conservative. Do not create task-specific memories.

Current query:
\begin{Verbatim}[breaklines=true, fontsize=\small]
{{CURRENT_QUERY}}
\end{Verbatim}

Trajectory:
\begin{Verbatim}[breaklines=true, fontsize=\small]
{{TRAJECTORY}}
\end{Verbatim}

Final answer:
\begin{Verbatim}[breaklines=true, fontsize=\small]
{{FINAL_ANSWER}}
\end{Verbatim}

Task feedback:
\begin{Verbatim}[breaklines=true, fontsize=\small]
{{TASK_FEEDBACK}}
\end{Verbatim}

Retrieved insights:
\begin{Verbatim}[breaklines=true, fontsize=\small]
{{RETRIEVED_INSIGHTS_JSON}}
\end{Verbatim}

Return only valid JSON in this format:

\begin{Verbatim}[breaklines=true, fontsize=\small]
{
  "insight_edits": [
    {
      "insight_id": "...",
      "operation": "KEEP|ARCHIVE|REVISE",
      "revised_insight_text": null
    }
  ],
  "add_new_insight": {
    "decision": false,
    "new_insight_text": null
  }
}
\end{Verbatim}

\end{tcolorbox}

\end{document}